\documentclass[sn-mathphys,Numbered]{sn-jnl}

\usepackage{graphicx}%
\usepackage{multirow}%
\usepackage{amsmath,amssymb,amsfonts}%
\usepackage{amsthm}%
\usepackage{mathrsfs}%
\usepackage[title]{appendix}%
\usepackage{xcolor}%
\usepackage{textcomp}%
\usepackage{manyfoot}%
\usepackage{booktabs}%
\usepackage{algorithm}%
\usepackage{algorithmicx}%
\usepackage{algpseudocode}%
\usepackage{listings}%
\usepackage{subfig}
\usepackage{amsmath}
\usepackage{amssymb}
\usepackage{booktabs}
\usepackage{tabularx}
\usepackage{ifsym}
 \usepackage{mathptmx}      
\usepackage{amsmath}
\usepackage{amssymb}

\theoremstyle{thmstyleone}%
\theoremstyle{thmstyletwo}%

\theoremstyle{thmstylethree}%

\begin{document}

\title[Article Title]{Deep Neural Networks Fused with Textures for Image Classification}

\author*[1]{\fnm{Asish } \sur{Bera}}\email{asish.bera@pilani.bits-pilani.ac.in}
\author[2]{\fnm{Debotosh } \sur{Bhattacharjee}}
\author[2]{\fnm{Mita } \sur{Nasipuri}}

\affil*[1]{\orgdiv{Department of Computer Science and Information Systems}, \orgname{BITS Pilani}, \orgaddress{\street{Rajasthan},  \country{India}}}

\affil[2]{\orgdiv{Department of Computer Science and Engineering}, \orgname{Jadavpur University}, \orgaddress{\city{Kolkata}, \state{WB}, \country{India}}}

\abstract{Fine-grained image classification (FGIC) is a challenging task in computer vision for  due to small visual differences among inter-subcategories, but, large intra-class variations. Deep learning methods have achieved remarkable success in solving FGIC. In this paper, we propose a fusion approach to address FGIC by combining  global texture  with local patch-based information. The first pipeline extracts deep features from various fixed-size non-overlapping patches and encodes features by sequential modeling using the long short-term memory (LSTM). Another path computes image-level textures at multiple scales using the local binary patterns (LBP). The advantages of both streams are integrated to represent an efficient feature vector for image classification. The method is tested on eight datasets representing the human faces, skin lesions, food dishes, marine lives, etc. using four standard backbone CNNs. Our method has attained better classification accuracy over existing methods with notable margins. }




\maketitle

\section{Introduction}\label{sec1}

Fine-grained image classification (FGIC) is a  challenging problem in computer vision since the past decades \cite{wei2019deep}. It discriminates smaller visual variations among various sub-categories of objects like human faces, flowers, foods, etc. The convolutional neural networks (CNNs) have achieved  high performance in FGIC. The CNNs represent object's shape, texture, and other correlated information in the feature space. In addition with global image-level description, object-parts relation, and local patch information  have shown their efficacy by mining finer details to solve FGIC. Many works have been devised leveraging attention mechanism 
 \cite{liu2021learning}, \cite{bera2021attend}; context encoding \cite{ge2019weakly}, \cite{behera2021context};   
erasing data augmentation \cite{zhong2020random}, \cite{bera2023fine}, and others. Many works avoid bounding-box annotations to localize essential image-regions using weakly supervised part selection \cite{liu2021cross}. Thus, defining region-based  descriptors is a key aspect to enhance FGIC performance.  

In another direction, the local binary patterns (LBP) \cite{ojala2002multiresolution} have achieved significant success in describing textural features from human faces, and other image categories \cite{bi20212d}. LBP is a non-parametric texture descriptor, extracted from a grayscale image. It encodes the differences between a pixel and its neighborhood pixels localized in a rectangular grid \textit{e.g.}, 3$\times$3, etc. Here, both  the textural and deep features are fused to formulate an efficient feature vector for image classification. 
This work proposes a method namely, \textbf{D}eep (Neural) \textbf{N}etworks fused with \textbf{T}extures (DNT) to explore its aptness for FGIC.  The first path extracts deep feature map using a base CNN. Then, the high-level deep feature maps is pooled through a set of non-overlapping patches. Next, a global average pooling (GAP) layer is applied to summarize the features followed by patch-encoding using the long short-term memory (LSTM). The other path computes the histograms of LBPs as  local texture-based feature descriptors. Finally, these two sets of features are mixed prior to a classification layer. 
 We have experimented on eight small-scale image datasets (1k-15k), representing a wide variations in the object's shape, color, background,  texture, etc. The datasets includes human faces with age-variations \cite{panis2016overview}, \cite{panis2014overview}; hand shapes/palmprint \cite{charfi2016local}, \cite{mukherjee2022human},  skin lesions \cite{olayah2023ai}, natural objects like flowers, underwater sea-lives; and food-dishes of India \cite{pandey2017foodnet} and Thailand \cite{termritthikun2017accuracy}.  . Currently, this extended work contains new results tested on two more datasets representing hand-shape and skin lesions, including the results of previous work. The contributions of this paper are as follows: 
 \begin{itemize}
     \item The deep features  and local binary patterns are fused for image recognition.
     \item  The method achieves satisfactory accuracy on eight image datasets representing the human faces, hand, skin lesions, food dishes, and natural object categories.  
 \end{itemize}

The rest of this paper is organized as follows: Section \ref{rel_work} summarizes related works, and Section \ref{proposed} describes the proposed method. The experimental results are discussed in Section \ref{experiments}, followed by the conclusion in Section \ref{conclusion}.

\section{Related Works} \label{rel_work} 
Human faces, food items, and other objects  (\textit{e.g.}, flowers, marine lives, etc.) recognition is  a challenging FGIC task. Apart from global feature descriptor rendered from a full-image, patch-descriptors have attained remarkable progress using deep learning. Part-based methods focusing on local descriptions and semantic correlations are integrated \cite{bera2021attend}. In this direction, multi-scale region proposals, and fixed-size patches have attained much research attention. In \cite{behera2021context}, multi-scale region features are encoded via the LSTM units. In \cite{ge2019weakly}, mask-RCNN is employed to localize discriminative regions. Several approaches have explored attention mechanism to improve FGIC performance including food recognition \cite{bera2021attend}. Few methods have proposed an ensemble of various CNNs,  fusion of two or more sub-networks for performance gain \cite{pandey2017foodnet}. Recently, vision transformers (ViT) have embedded non-overlapping patches with multi-head self-attention module \cite{dosovitskiy2020image}. The performance can be boosted by the random erasing data augmentation for multi-scale patch-based feature representation \cite{bera2023fine}.  Various food-dishes classification is discussed in \cite{tiankaew2018food}, \cite{lim2021explainable}, \cite{arslan2021fine}. Researchers have summarized a comprehensive study on FGIC \cite{wei2019deep}, and food recognition \cite{min2019survey}. The Forward Step-wise Uncertainty-Aware Model Selection  has described an deep learning based ensemble method for food-dishes classification \cite{aguilar2022uncertainty}. Underwater object detection, segmentation,  and classification  is a challenging research area in computer vision. Studies on the marine lives detection using deep learning techniques are summarized in \cite{xu2023systematic}, \cite{wang2023deep}. Also, early diagnosis of skin cancer  from skin lesions using hybrid models CNN-ANN and CNN-RF is presented  \cite{olayah2023ai}. A survey in this direction has been studied in \cite{hasan2023survey}. Several benchmark public datasets are developed by the International Skin Imaging Collaboration (ISIC)  for  detection and classification of skin cancer, melanoma, and lesions using dermoscopy images \cite{codella2018skin},  \cite{cassidy2022analysis}, \cite{li2018skin}. Marine-life classification  using transfer learning based on  pre-trained CNNs is described \cite{liu2019real}. A video dataset on underwater marine animals of six categories is proposed with baseline results \cite{pedersen2019detection}.

Before the deep learning era, different local shape and texture information including the LBP, geometric properties, shape profiles, bag of words, Scale Invariant Feature Transform (SIFT), colors, and many more have been described in the literature \cite{bera2017human}, \cite{pietikainen2015two}, \cite{charfi2016local}, \cite{bera2015fusion},
\cite{bera2021spoofing}, \cite{hasan2023survey}, \cite{olayah2023ai}, 
\cite{bera2014person}, \cite{barata2013two}, \cite{bera2021two}. Most of these conventional feature descriptors are used for recognizing the human faces, emotions, hand-shape, palmprint, skin lesions, and other biometric modality and object categories.  A significant number of works on face recognition have computed local textures using the LBP family and others. A classical gray-scale, rotation invariant, and uniform LBP at circular neighborhoods is introduced in \cite{ojala2002multiresolution}.  Recently, a learning 2D co-occurrence LBP  is described to attain scale invariance for image recognition \cite{bi20212d}.  Deep architectures have been developed underlying the LBP for textural feature extraction for face recognition \cite{xi2016local}. Empirical model with local binary convolution layers is explored \cite{juefei2017local}. Weighted-LBP emphasizes more importance on regions which are more influenced by aging effects \cite{zhou2018age}. Multi-scale LBP and SIFT descriptors are used for multi-feature discriminant analysis in \cite{li2011discriminative}. The  VGG-Face model is used for extracting  the features for face recognition \cite{moustafa2020age}. A journey of LBP since the past two decades is sketched in \cite{pietikainen2015two}.
With a brief study, this paper explores a combination of deep features  using  CNN and texture features using LBP for  classifying images with diverse categories.

\section{{Proposed Method: Deep Networks fused with Textures}} \label{proposed}
\begin{figure}
\centering
\includegraphics[width= 0.97 \textwidth, height=5.0 cm ]{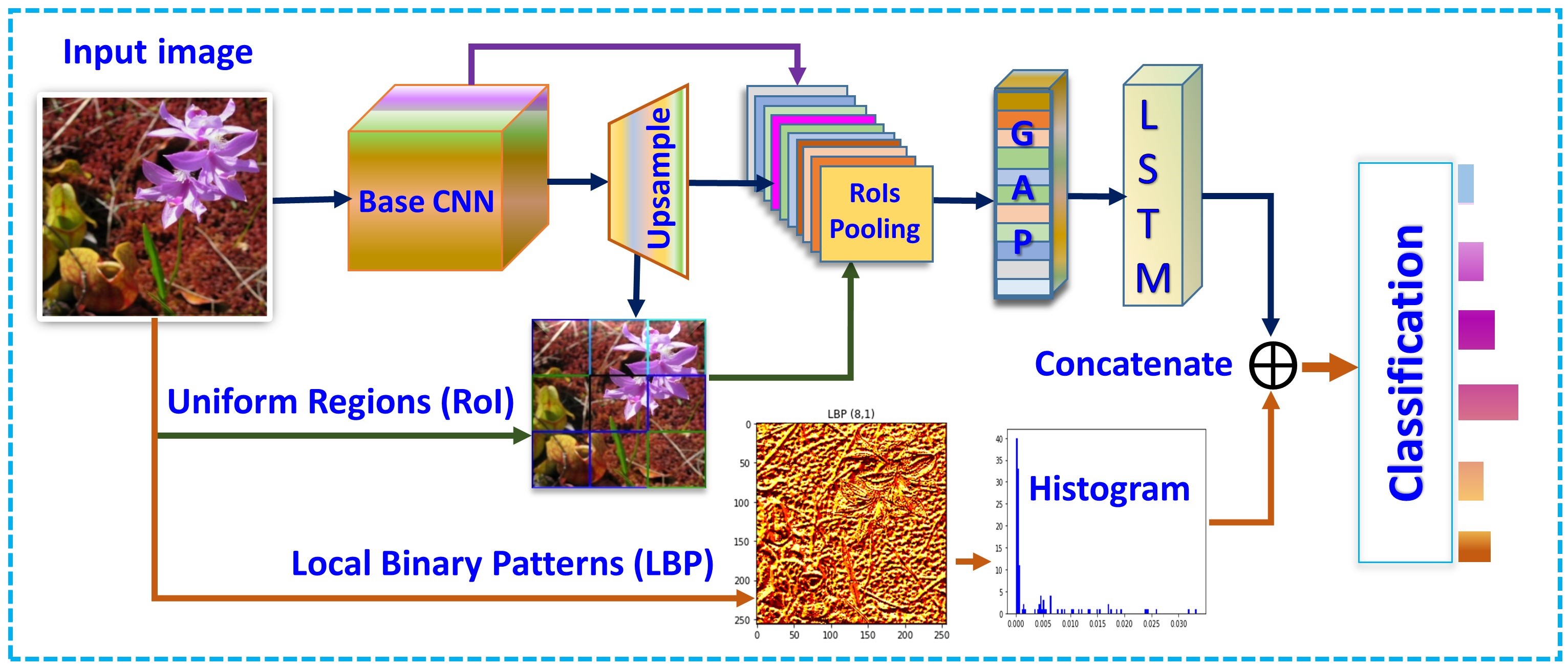}
\caption{Proposed method (DNT) fuses deep features and texture descriptors using local binary patterns (LBP) for fine-grained image classification. } \label{model}
\end{figure}

Proposed DNT is a two-stream deep model (Fig. \ref{model}). Firstly, it emphasizes the features via patches and LSTM. Then, it combines multiple LBP. Lastly, both paths are fused.
\subsection{Convolutional Feature Representation}
An input color image with its class-label $I_l$ $\in$ $\mathbb{R}^{h\times w\times 3}$ is fed into a base CNN, such as DenseNet-121, etc.  
A CNN, say  ${N}$  extracts high-level feature map ${F}$ $\in$ $\mathbb{R}^{h\times w\times c}$ where $h$, $w$, and $c$ denote the height, width, and channels, respectively. Simply, we denote ${N}(I_l, \theta)\rightarrow \textit{F}$ to compute deep features, where image $I_l$ is provided with its class-label $l$, and $\theta$ represents the learning parameters of ${N}$. The feature map $\textit{F}$ from the last convolutional layer of  base  ${N}$ is extracted to develop the proposed model by including other functional modules. 
\subsection{Patch Encoding}
The region proposals ($\textit{D}$) are generated as non-overlapped  uniform (same size) patches from $I_l$. The  resulting number of regions is $e={(h\times w)}/a^2$, where $a\times a$ is spatial size of a rectangular patch $d$. A set $\textit{D}=\{d_1,d_2,..., d_e | I_l\}$ of $e$ patches are pooled from feature map ${F}$ which is spatially upsampled to $h'\times w'\times c$ size prior to pooling. The  patches represent fine-details and local contexts  which are important for subtle discrimination in FGIC. The bilinear pooling is applied to compute features from every patch of  size $h_1\times w_1\times c$. Next, the global average pooling (GAP) is applied to summarize the mean features of  \textit{D}. It downsamples the spatial dimension at patch-level  to $1\times 1 \times c$. The resulting feature map is ${F}_1 $. To  learn effectiveness of patches, a single layer fully-gated LSTM \cite{hochreiter1997long} is applied  to learn long-term dependencies via the hidden states. The  encoded feature vector is denoted as $F_2\in \mathbb{R}^{v\times 1}$, defined in (Eq. \ref{eq1}).
\begin{equation} \label{eq1}
\centering
\begin{split}
{F}={N}(I_l, \theta);  \hspace{2 mm} {F}_1=N\Bigl(\textit{D}, {GAP}({F}), \theta_1 \Bigl); \hspace{2 mm}  {F}_2= N\Bigl(D, LSTM ({F}_1), \theta_2 \Bigl)  \\
\end{split}
\end{equation}
\subsection{Textures Representation using Local Binary Patterns}
The local binary pattern ($LBP$) is a monotonic grayscale-invariant local descriptor which computes spatial textures. The histogram of $LBP$ labels is considered as a feature vector. Here, the uniform value of $LBP_{P, R}$ is extracted as texture descriptor at global image-level. $P$ defines total number of sampled neighbors, and $R$ represents the radius of circular neighborhood. 
\begin{equation} \label{eq2}
\centering
{LBP}_{P, R}= {\sum_{i=0} ^ {P-1} q(p_i -p_c).2^i}; \hspace{ 1 cm}  q(p_i -p_c)=\begin{cases}
    1, & \text{if $(p_i -p_c) \geq 0$}\\
    0, & \text{otherwise}
  \end{cases} 
\end{equation}
\noindent Here, $p_c$ denotes grayscale value of center-pixel of a local window, and $p_i$ represents value of corresponding neighbor pixel of $p_c$, and $q(.)$ is an indicator function. The histograms of multiple neighborhoods are combined to improve the effectiveness of texture patterns. The descriptor ${F_3}$ is defined as
\begin{equation} \label{eq3}
\centering
{F_3} = \overset{P, R}{\underset{i=j=1}{\Bigm\Vert}} LBP(I)_{P_i, R_j}; {F_{final}} = N(F_2 {\Bigm\Vert} F_3, \theta_f); \hspace{ 2 mm}  \bar{l}=\textit{softmax}({F}_{final}) ; \hspace{ 2 mm} \bar{l} \in  \mathbb{R}^{Y\times 1}
\end{equation}
\noindent where, $\Bigm\Vert$ denotes concatenation operator. The neighborhood spatial structures of $P=8, 16$ and $R=1, 2$ combinations are considered, shown in top-row of Fig. \ref{erasing}. The dimension of combined image-level texture vector is $4\times 256=1024$. However, other higher values can also be computed according to Eq. \ref{eq2}-\ref{eq3}.

\subsubsection{Fusion} 
Finally, ${F_2}$ and ${F_3}$ are concatenated to produce a mixed feature vector $F_{final}$ which is fed to a \textit{softmax} layer for generating an output probability vector implying each predicted class-label $\bar{l}$  corresponds to  actual-label $l \in Y$ from a set of classes $Y$. 

\subsection{Random Region Erasing Image Augmentation}
\begin{figure}
\subfloat{\includegraphics[width=0.2\textwidth]{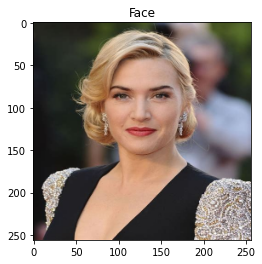}} \hfill
\subfloat{\includegraphics[width=0.2\textwidth]{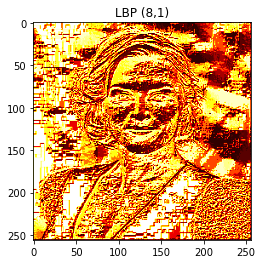}} \hfill
\subfloat {\includegraphics[width=0.2\textwidth]{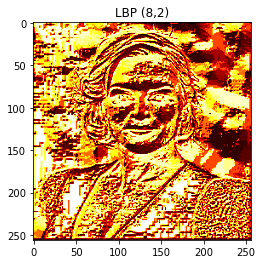}} \hfill
\subfloat{\includegraphics[width=0.2\textwidth]{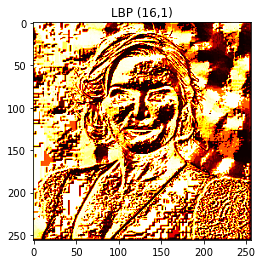}} \hfill
\subfloat{\includegraphics[width=0.2\textwidth]{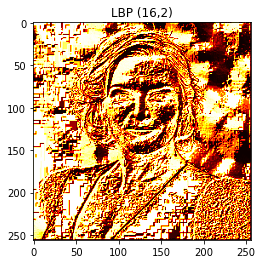}} \hfill

\subfloat{\includegraphics[width=0.16\textwidth]{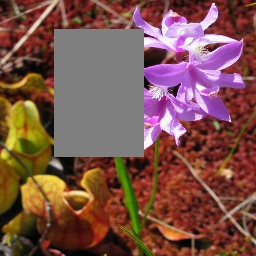}} \hfill
\subfloat{\includegraphics[width=0.16\textwidth]{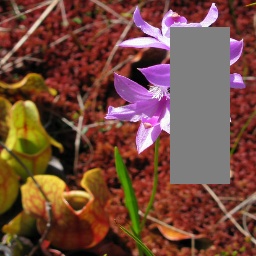}} \hfill
\subfloat{\includegraphics[width=0.16\textwidth]{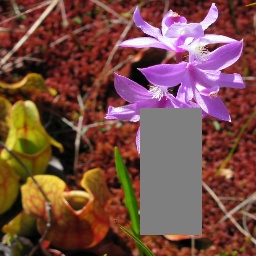}} \hfill
\subfloat{\includegraphics[width=0.16\textwidth]{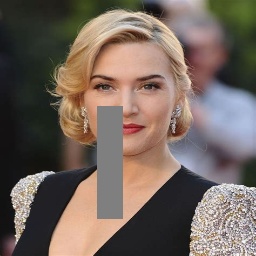}} \hfill
\subfloat{\includegraphics[width=0.16\textwidth]{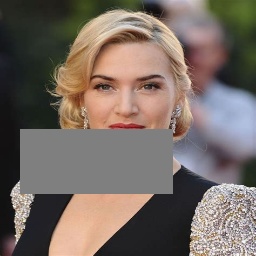}} \hfill
\subfloat{\includegraphics[width=0.16\textwidth]{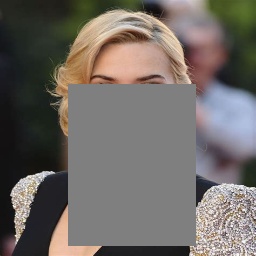}} \hfill
\caption{Top-row: LBP of various neighborhoods (P, R): (8,1), (8,2),  (16,1), and (16,2). \\ Bottom-row: Random erasing data augmentation on flower and celebrity-face images. }
\label{erasing}
\end{figure}

Several image augmentation methods are used \textit{e.g.}, translation, rotation, scaling, random erasing  \cite{zhong2020random}, \cite{bera2023fine}, etc. Here, random erasing at global image-level is applied along with general data augmentations. It randomly selects a rectangular region $I_E$ in  $I$, and erases pixels inside $I_E$ with random values within [0, 255]. The height and width of $I_E$ are randomly chosen on-the-fly within [0.2, 0.8] scale, and pixels are erased with  value 127. Examples are shown in the bottom-row of Fig. \ref{erasing}. 
\section{Experimental Results and Discussion} \label{experiments}

Firstly, the datasets are summarised, followed by the implementation details. The performance evaluation and comparison with state-of-the-arts are discussed next. 

\subsection{Dataset Summary} 
\begin{figure}
\centering
\includegraphics[width= 0.24 \textwidth, height= 3cm] {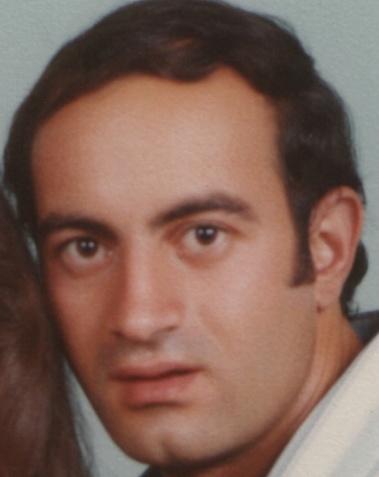} \hfill
\includegraphics[width= 0.24\textwidth, height= 3cm] {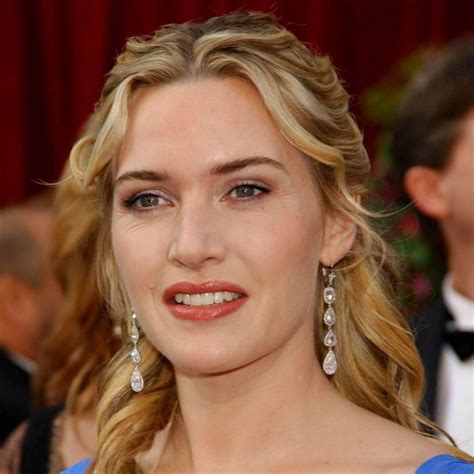} \hfill 
\includegraphics[width= 0.24\textwidth, height= 3cm] {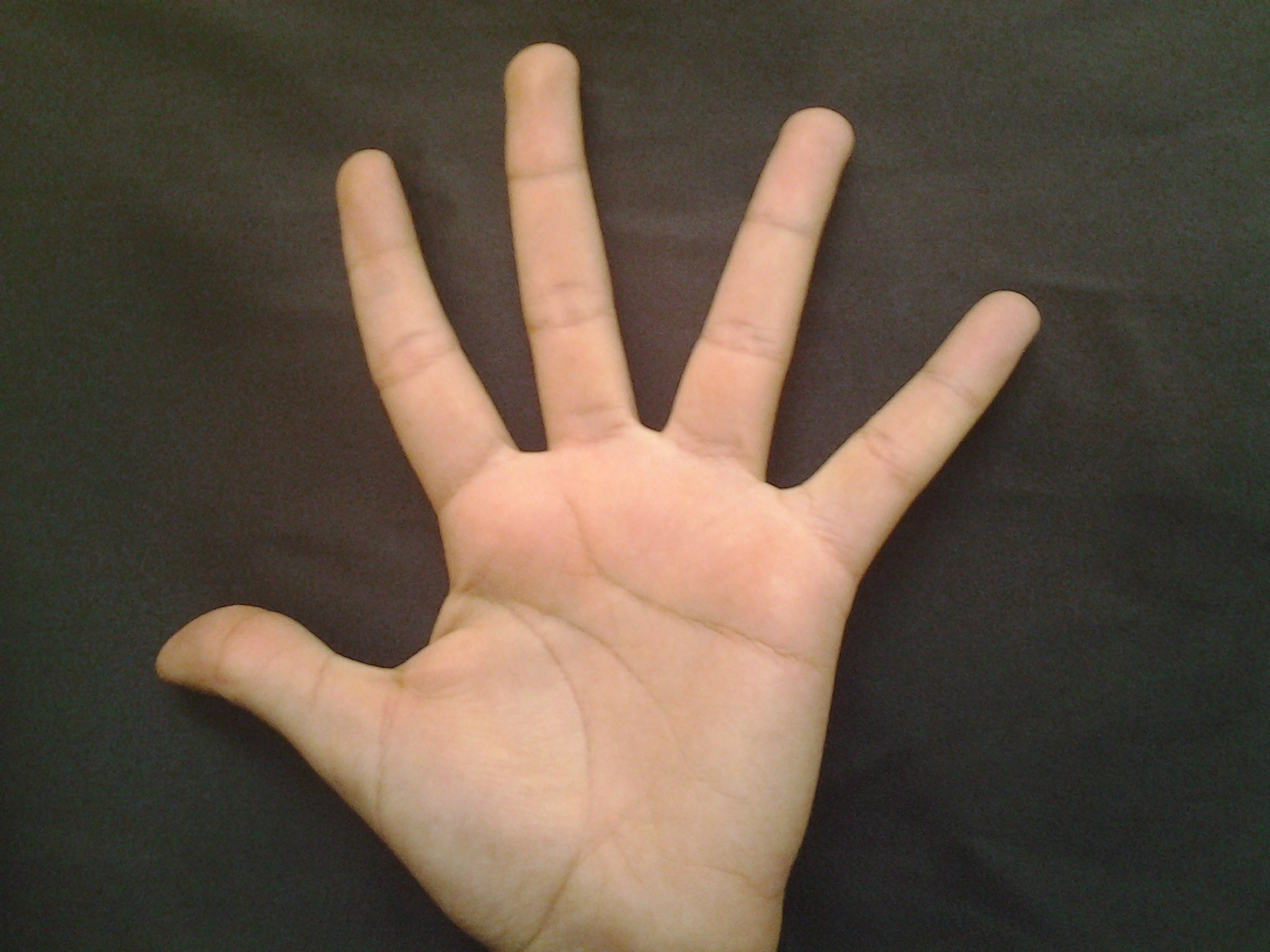} \hfill
\includegraphics[width= 0.24\textwidth, height= 3cm] {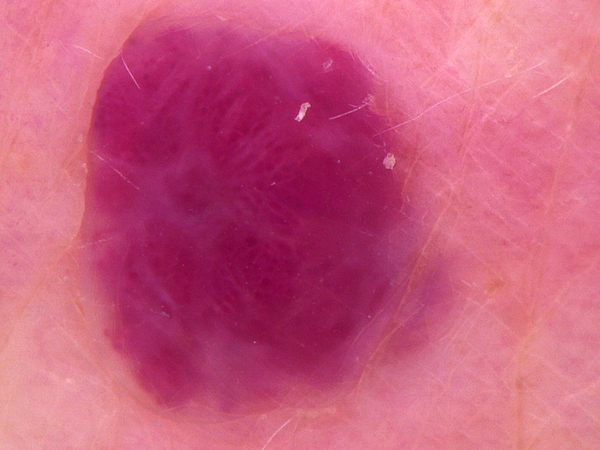}

\includegraphics[width= 0.24 \textwidth, height= 3 cm] {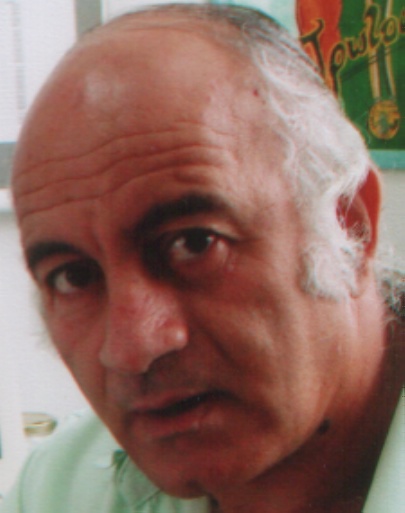} \hfill
\includegraphics[width= 0.24 \textwidth, height= 3 cm] {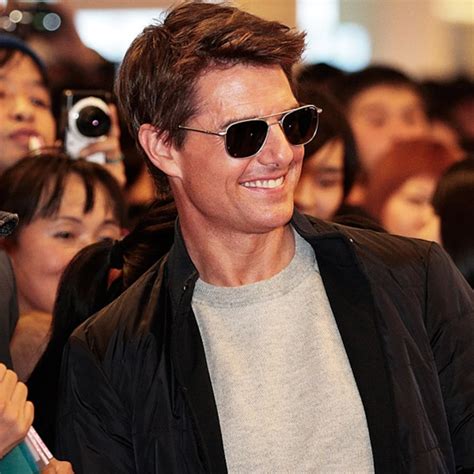} \hfill 
\includegraphics[width= 0.24\textwidth, height= 3 cm] {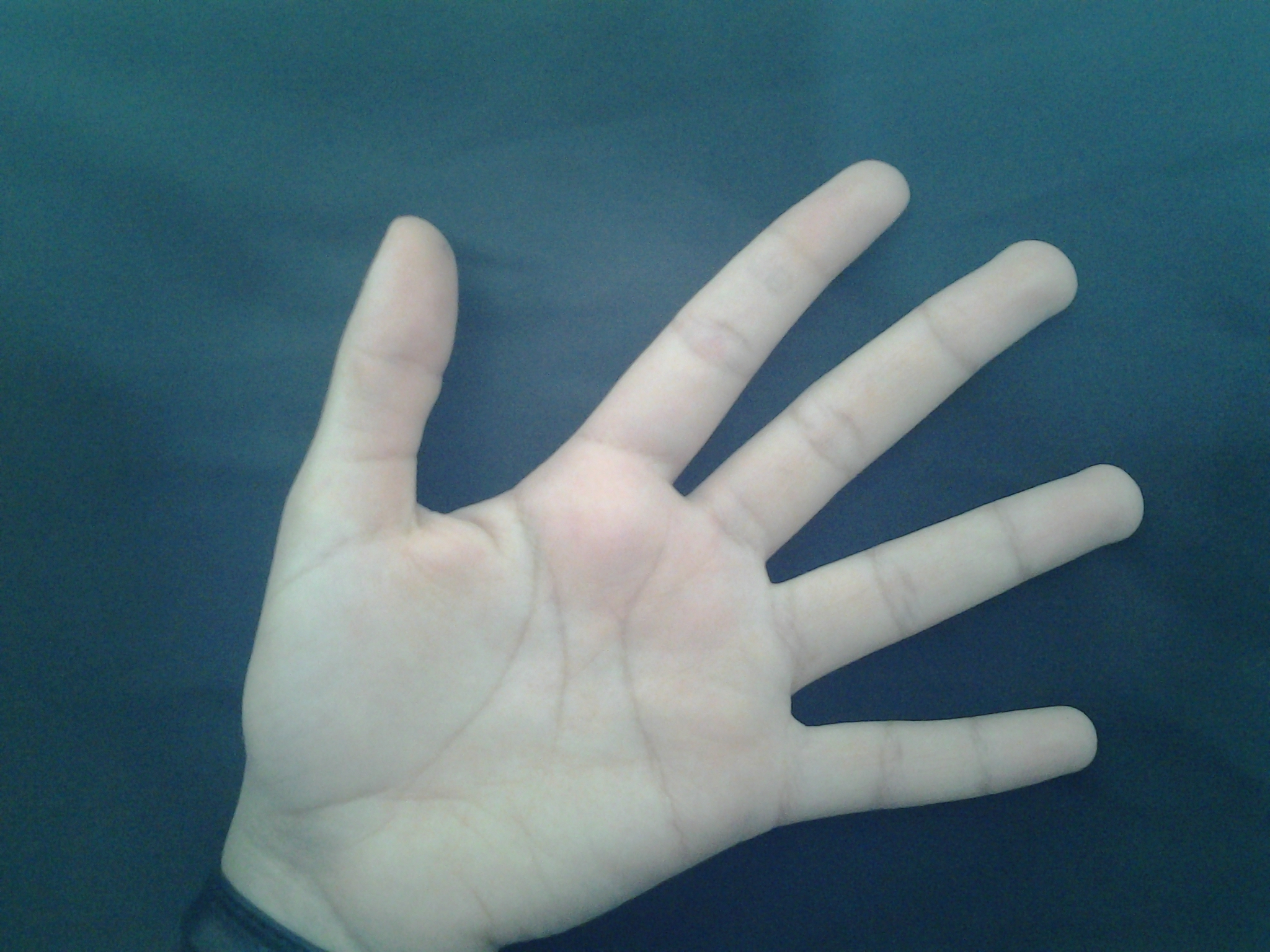} \hfill
\includegraphics[width= 0.24\textwidth, height= 3 cm] {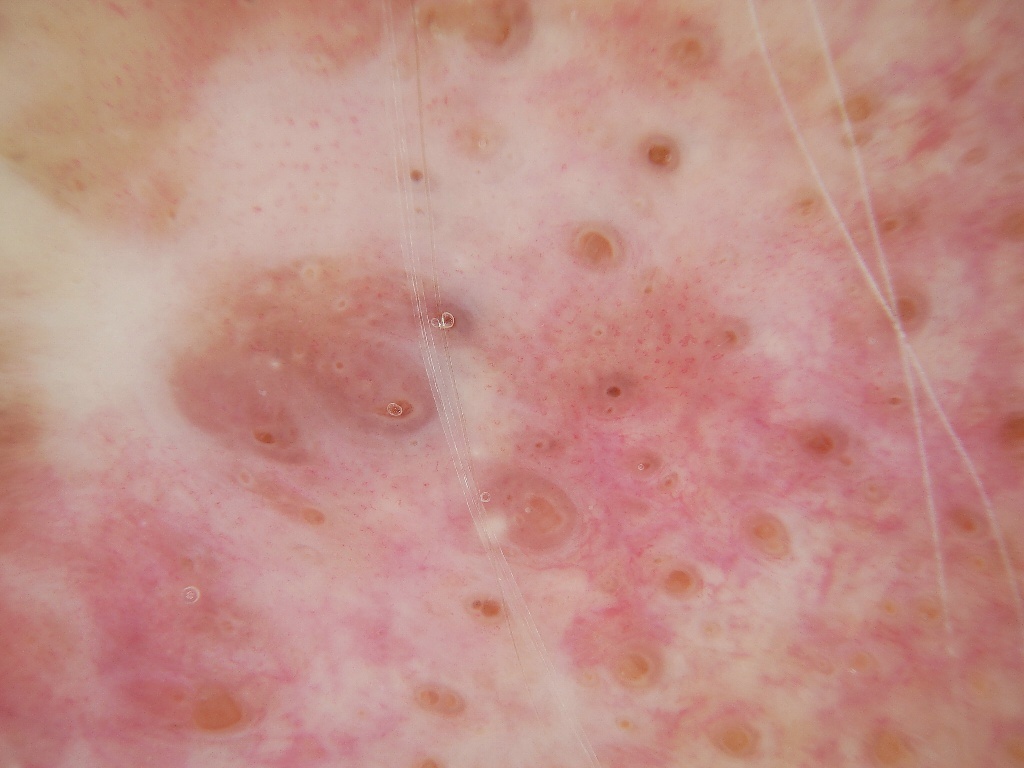} 

\caption{Dataset samples are shown column-wise: human faces of FG-Net and celebrity, hand shape, and ISIC skin lesions.}
\label{samples}
\end{figure}

\begin{figure}
\centering

\includegraphics[width= 0.24 \textwidth, height= 3.0 cm] {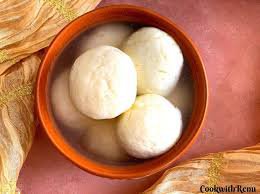} \hfill
\includegraphics[width= 0.24 \textwidth, height= 3.0 cm] {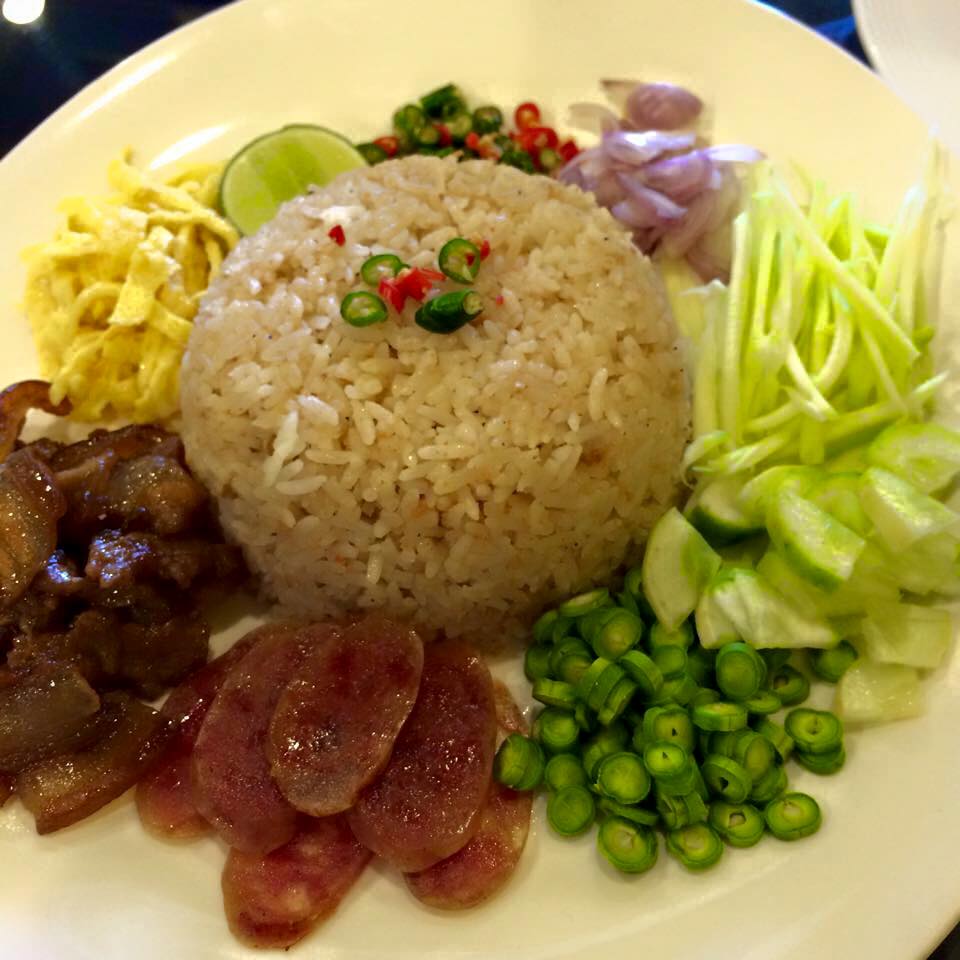} \hfill 
\includegraphics[width= 0.24 \textwidth, height= 3.0 cm ]{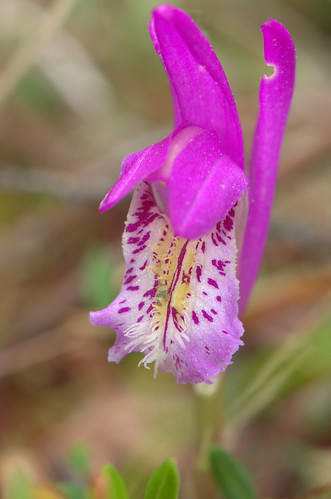} \hfill 
\includegraphics[width= 0.24 \textwidth, height= 3.0 cm] {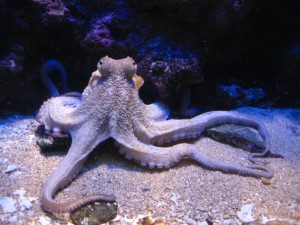 }

\includegraphics[width= 0.24 \textwidth, height= 3.0 cm] {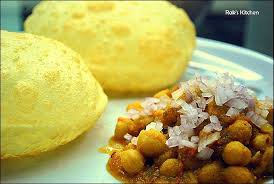} \hfill
\includegraphics[width= 0.24 \textwidth, height= 3.0 cm] {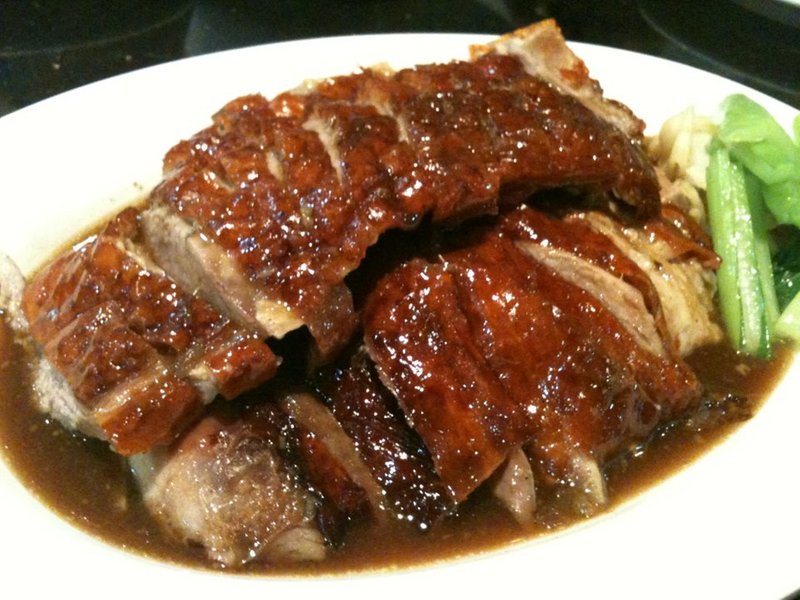} \hfill 
\includegraphics[width= 0.24\textwidth, height= 3.0 cm] {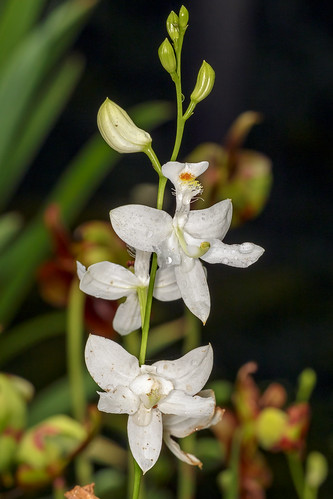} \hfill
\includegraphics[width= 0.24 \textwidth, height= 3.0 cm] {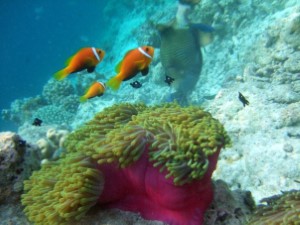 }

\caption{Dataset samples are shown column-wise: food-dishes of India and Thailand,  natural objects representing flower and  marine-lives.}
\label{samples2}
\end{figure}

Proposed DNT is evaluated on eight datasets representing the human faces, hand-shapes, food dishes, flowers,   marine-lives, and skin lesions. A well-known age-invariant human face dataset, FG-Net contains 1002 images of 82 persons with ages from  0 to 69 years \cite{panis2014overview}. A touch-less hand database, called REgim Sfax Tunisia (REST), is mainly used for palmprint (biometric) recognition using local texture and shape descriptors \cite{charfi2016local}. A subset of REST dataset containing at least 5 left-hand images (2 images for testing and remaining 3 or more for training per class) of 179 individuals each is used in our work. The datasets  comprised with 80 Indian dishes  \cite{pandey2017foodnet} and 50 Thailand dishes \cite{termritthikun2017accuracy} are tested. Remaining 4 datasets \textit{i.e.}, the celebrity faces \footnote{https://www.kaggle.com/datasets/vishesh1412/celebrity-face-image-dataset}, flowers \footnote{https://www.kaggle.com/datasets/alxmamaev/flowers-recognition}, marine animals \footnote{https://www.kaggle.com/datasets/vencerlanz09/sea-animals-image-dataste}, and skin lesions \footnote{https://www.kaggle.com/datasets/nodoubttome/skin-cancer9-classesisic}  are collected from the Kaggle repository. The images are randomly divided into decent train-test sets, detailed in Table \ref{exp1}. Dataset samples are shown in Fig. \ref{samples}-\ref{samples2}. The top-1 accuracy (\%) is evaluated for assessment and the model parameters are estimated in millions (M). 
\subsection {Implementation} \label{implemnt}
The  DenseNet-121, DenseNet-201, ResNet-50, and MobileNet-v2 backbone CNNs are used for deep feature extraction, and fine-tuned on the target datasets. Pre-trained ImageNet weights are used to initialize base CNNs with input image-size 256$\times$256. Random region erasing, rotation ($\pm$25 degrees), scaling ($\pm$0.25), and  cropping   224$\times$224 image-size are followed for data augmentation. The output feature map (\textit{e.g.}, 7$\times$7$\times$c) is upsampled to 48$\times$48$\times$c for pooling of $4\times4$ patches, and the value of output channels (c) varies according to CNN architectures, \textit{e.g.}, $c=1024$ for DenseNet-121. Uniform patch-size is 12$\times$12 pixels  to generate 16 patches. The feature size of LSTM's hidden layers is 1024, and concatenated with LBP of same size. The final feature vector $c=2048$ is fed to the \textit{softmax} layer for classification. Batch normalization and drop-out rate is 0.2 is applied to ease over-fitting.  The Stochastic Gradient Descent (SGD) optimizer is used to optimize the categorical cross-entropy loss  with an initial learning rate of $10^{-3}$ and divided by 10 after 100 epochs. The DNT model is trained for 200 epochs with a mini-batch size of 8 using  8 GB Tesla M10 GPU, and scripted in  Python. 
\begin{table}
\centering
\caption{Dataset summary and  test results using 3$\times$3 patches and 256$\times$256 LBP }\label{exp1}
\begin{tabular}{|c|c|c|c|c|c|c|c|}
\hline
Dataset & Class & Train & Test &  DenseNet121  & ResNet50 &  DenseNet201   & MobileNetv2   \\
\hline
FG-Net & 82 & 827 & 175 & 52.38 & 48.80  & 57.73  &52.38 \\
Celebrity & 17 &1190 & 510 & 94.24 & 89.28 & 95.04  & 92.85 \\
Indian Food & 80 &2400   & 1600  & 72.18 & 68.87  & 73.31   & 69.62  \\
Thai Food & 50 &14172  & 1600  & 92.31  & 90.18  & 92.50   & 89.93 \\
Flower &   6 & 2972   & 1400  & 96.85   & 95.78 &96.71  & 96.07  \\
Sea-life  & 18 & 5823&3636 &90.36 &89.10 &91.05  &90.09\\
REST-Left Hand & 179 & 616 &358 &79.73 &81.25 &80.96 &77.27\\

ISIC Skin Cancer & 7 & 1491 & 675 & 78.86 &77.23 & 79.45&73.06 \\ \hline
\multicolumn{4}{|c|}{{Model Parameters (Millions) }}  & 10.2 & 29.0  & 23.5  &6.1\\
\hline
\end{tabular} 
\end{table}
\subsection{Result Analysis and Performance Comparison}
The test results with $3\times3$ patches and two LBP structures \textit{i.e.}, (8, 1) and (8, 2) with a total 512 textures are  given in Table \ref{exp1}. The feature size of LSTM's hidden unit is 512, and after concatenating with histograms of LBP, size of final feature map is 1024. The last-row estimates the parameters (Millions) of various models. The accuracy (\%) is very decent, except age-invariant face recognition (AIFR) \textit{i.e.}, FG-Net.  Many existing methods have experimented on FG-Net dataset for AIFR by following leave-one-person out strategy \cite{moustafa2020age}, \cite{zhao2020towards}. In our set-up, FG-Net test-set includes at least one unseen image per person by splitting 1002 samples into train-test (83:17) sets. Here, we have tested this challenging dataset for FGIC rather than AIFR. Hence, DNT is not directly comparable with existing methods. However,  DNT attains better results (Table \ref{exp1} and Table \ref{tab3}) than NTCA (48.96\%) \cite{bouchaffra2014nonlinear}, and other works on AIFR \cite{panis2016overview}. 

\begin{figure}
\includegraphics[width =0.48\textwidth]{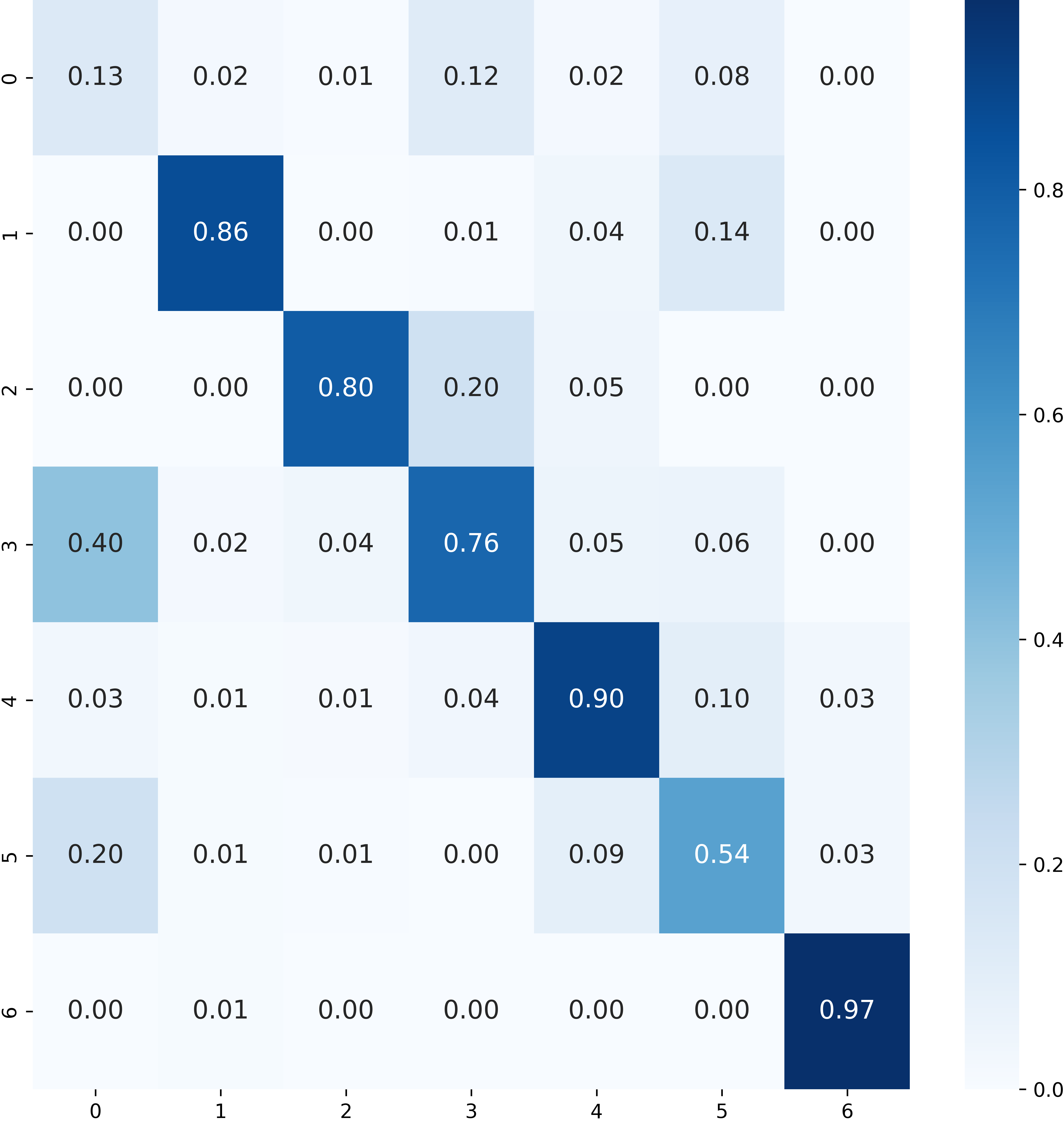} \hfill
\includegraphics[width =0.48\textwidth]{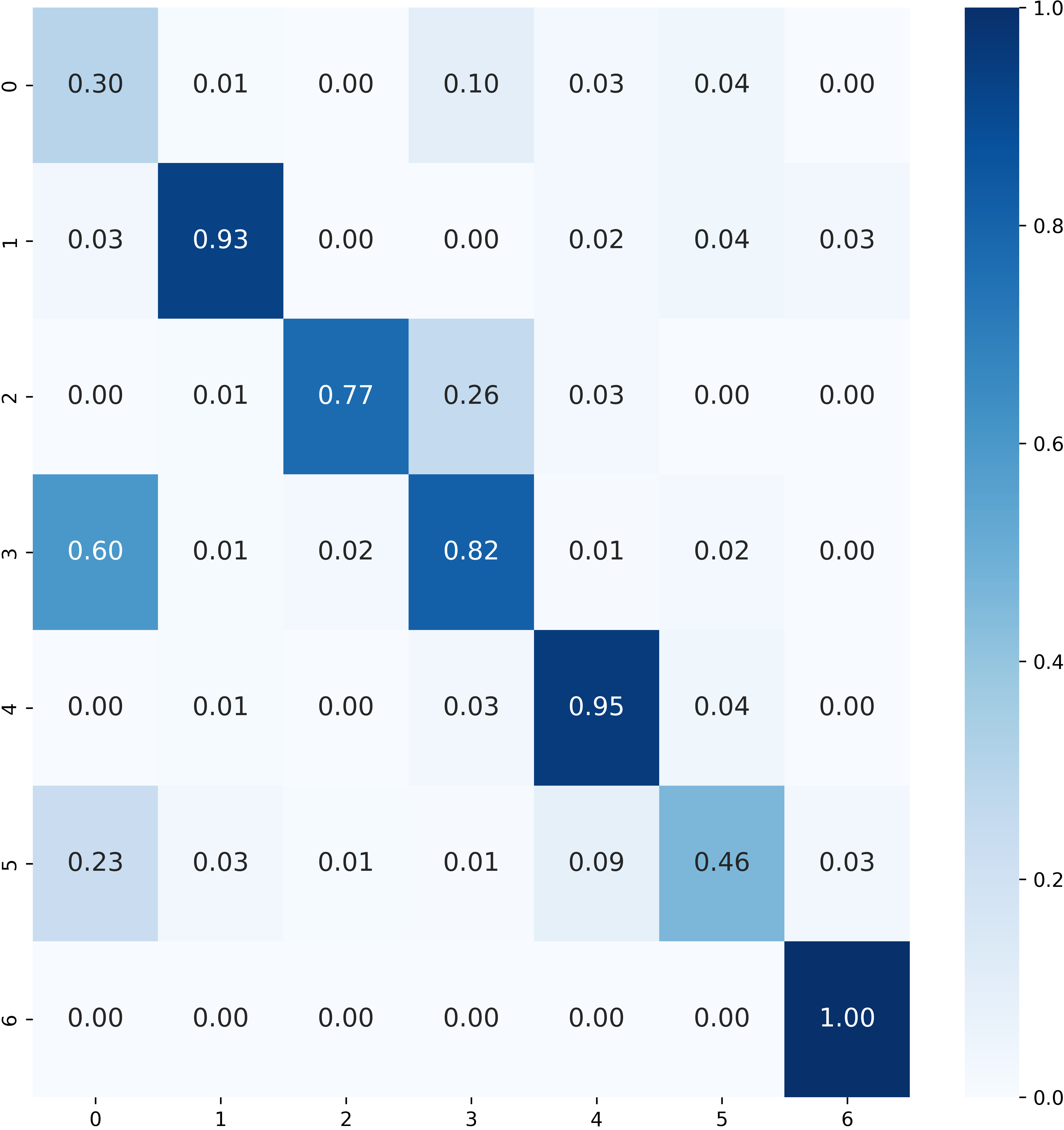}
\caption{Confusion matrix on ISIC skin cancer dataset using DNT with $4\times4$ patches and $2\times1024$ LBP based on ResNet-50 (left)  and DenseNet-201 (right) backbone CNNs.}
\label{cm}
\end{figure}

The REST dataset \cite{charfi2016local} is primarily tested for palmprint based biometric identification using traditional textures and more recently using CNNs. In this work, we have tested the left-hand images of REST dataset for classification using full hand-shapes and its class-labels only, \textit{i.e.}, without determining any region of interests for palmprint, or additional pre-processing stage. The proposed DNT has achieved 85.79\% top-1 accuracy using DenseNet-201. The precision is 90.0\% and recall is 86.0\%. On the contrary, the SIFT descriptors \cite{charfi2016local} reported 80.83\% palmprint identification success with the samples of 150 persons. Though, our DNT is not directly comparable with this existing work, yet, the gain of our method is significant on this dataset.
The ISIC 2017 dataset comprising with 2000 lesion images are classified into 3 categories, namely  Melanoma, Seborrheic keratosis and Nevu in \cite{li2018skin}. The classification accuracy is 85.7\% using Lesion Indexing Network (LIN). However, our method is not directly comparable with LIN \cite{li2018skin}. Our DNT has classified 7 skin  diseases namely actinic keratosis, basal cell carcinoma, melanoma, nevus, pigmented benign keratosis, squamous cell carcinoma, and vascular lesion. The accuracy is   { 81.10\% } using DenseNet-201 backbone. The confusion matrices are shown in Fig. \ref{cm}, representing the performance of DNT using ResNet-50 and DenseNet-201 base CNNs.

FoodNet presents classification of 50 Indian food-dishes  \cite{pandey2017foodnet}, and achieves 73.50\% accuracy using an ensemble method. It consists  100 images per class and 80\% images per class are used for training. We have used similar 80 dishes with 50 images per class, following 60:40 train-test ratio. DNT  achieves  80.75\%  and 74.75\%  accuracy using DenseNet-201 and ResNet-50, respectively (Table \ref{tab3}). We have tested on ThaiFood-50 \cite{termritthikun2017accuracy}, and the accuracy is 80.42\%. In \cite{termritthikun2018nu}, the accuracy is 83.07\% using ResNet-50. On the contrary, DNT attains 95.18\% using DenseNet-201, and 91.93\% by ResNet-50 (Table \ref{tab3}). 

The performance on other datasets are also high. However, to the best of our knowledge, no significant results have been reported on dataset, like sea-life. We have reported the results in the context of FGIC on these datasets for further research. 

\begin{figure}
\subfloat{\includegraphics[width =0.48\textwidth]{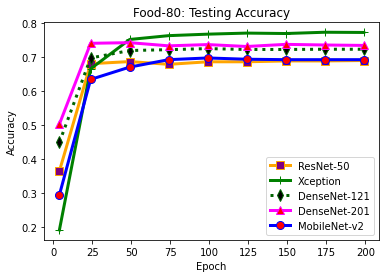}} \hfill
\subfloat{\includegraphics[width =0.48\textwidth] {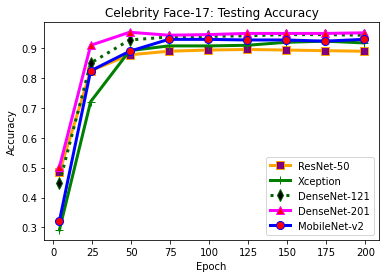}}
\caption{Test accuracy of various CNNs  using $3\times3$ patches and $2\times256$ LBP on Indian food-80 and celebrity faces datasets.   }
\label{val1}
\end{figure}

The overall classification performances of various CNNs  using $3\times3$ patches and $2\times256$ LBP on Indian food dishes and celebrity faces  are shown in Fig. \ref{val1}. It is evident that accuracy improvement is very small after 100 epochs. ResNet-50 and Xception are comparatively heavier models than the DenseNet family regarding the model parameters. Whereas, MobileNet-v2 is a lightweight model, yet,  very efficient for FGIC. 

Next, experiments on Indian food and flower show accuracy gain using $4\times4$ patches while other components are unaltered. This test is comprised with 16 patches, 512 LBP and 512 LSTM features. The results imply more patches improve accuracy (Table \ref{abl5}). 
\begin{table}
\centering
\caption{Performance of DNT using 16 patches and 512 LBP }\label{abl5}
\begin{tabular}{|c|c|c|c|c|}
\hline
Dataset  &  DenseNet-121  & ResNet-50 &  DenseNet-201   & MobileNet-v2   \\
\hline
Flower & 97.10 &94.78  &96.50   & 95.71  \\ 
Indian Food   &72.13  & 71.43   & 76.06   &  72.94 \\
\hline
Param (M)  & 10.3  & 28.8  & 23.3  &6.0\\
\hline
\end{tabular} 
\end{table}
\begin{table}
\centering
\caption{Performance  of DNT using 16 patches and 1024 LBP}\label{tab3}
\begin{tabular}{|c|c|c|c|c|}
\hline
Dataset  &  DenseNet-121  & ResNet-50 &  DenseNet-201   & MobileNet-v2   \\
\hline
FG-Net & 54.74  & 49.40  &   \textbf{55.95}  & 53.57 \\ 
Celebrity  & 95.43 & 92.06  &  \textbf{95.83}  & 90.87 \\ 
Indian Food   & 78.18 & 74.75  & \textbf{80.75}    & 76.31  \\ 
Thai Food   & 94.00 & 91.93  & \textbf{95.18}   & 92.75 \\ 
Flower &  97.50  & 97.14 & \textbf{98.00}  & 97.21\\ 
Sea-life  & 92.50  & 92.51 & \textbf{94.51}    & 92.34 \\ 
REST-Left Hand & 83.52  & 83.23& \textbf{85.79}    &80.14  \\ 
ISIC Skin Cancer & 80.50 &78.42 & \textbf{81.10} &77.52\\
\hline
Param (M)  & 15.8 & 36.5  & 30.8 & 12.1\\
\hline
\end{tabular} 
\end{table}

We have increased the number of LSTM's hidden states and  number of patches. This test is carried out with the fused features of 1024 textures (LBP), and 1024 LSTM units encoded from 16 patches. The results are reported in Table \ref{tab3}. Clearly, DenseNet-201 performs the best among four base CNNs, while other backbones produce satisfactory results. 

The significance of major components of DNT are tested, and the results are given in Table \ref{Abln1}. Particularly, the benefits of random erasing over general image augmentation, number of patches, textures (LBP), LSTM, and their further increment in the feature space are  investigated for performance improvement on two datasets using DenseNet-121 (DN-121). The ablative results justify the usefulness of essential components of the proposed DNT.

\begin{table}
\centering
\caption{Ablation study on proposed DNT using DenseNet-121 (DN-121) }\label{tab5}
\begin{tabular}{|c|c|c|c|}
\hline
{DenseNet-121 (DN-121) base CNN with key modules} & {Indian Food}  & {Sea-life} & Param\\
\hline
DN-121  + common image augmentation $  $ &  63.37 & 86.24   &7.0  \\
\hline
DN-121 + common + random erasing image augment $  $ &  67.25   & 88.43   &7.0\\
\hline
 DNT (DN-121) with 9 patches, and LBP (addition) &  71.49 & 89.23  & 10.2  \\
 \hline
\textbf{DNT} (DN-121)  with 16 patches, and without LBP &  {74.56}  & {90.28}  & {15.5} \\
\hline
\textbf{DNT} (DN-121) with 16 patches, and LBP (concatenation) &  \textbf{78.18}  & \textbf{92.50}  & {15.5}  \\ \hline
\end{tabular} \label{Abln1}

\end{table}

\vspace{0.5 cm}
\section{Conclusion}  \label{conclusion}
In this paper, we have presented a new work on image classification by fusing the deep features with local textures at image-level. The performance is evaluated  using four base CNNs on eight diverse FGIC datasets. We have achieved better results on these datasets compared to mentioned  existing works. In addition with conventional image augmentation, random region erasing data augmentation  improves the accuracy. The ablation study reflects the usefulness of important components. In future, we plan to develop a new model to improve the performance further and explore other fusion strategies for wider applicability on diverse datasets.  

\section*{Acknowledgement}
We  thank to the  Reviewers  to improve this paper. We thank to the repositories for the datasets used in this work. We are thankful to BITS Pilani,  Pilani Campus, Rajasthan, India, for providing necessary infrastructure and Research Initiation Grant (RIG) to carry out this work. 

\vspace{0.5 cm}
\bibliography{sn-bibliography}

\end{document}